\pdfoutput=1
\documentclass[11pt]{article}

\usepackage{acl}

\usepackage{times}
\usepackage{latexsym}
\usepackage{url}
\usepackage{hyperref}
\usepackage[all]{hypcap}
\usepackage{graphicx}
\usepackage{amsmath}
\usepackage{amssymb}
\usepackage{xspace}
\usepackage{fixfoot}
\usepackage{multirow}
\usepackage{ marvosym }
\usepackage{calc}
\usepackage{booktabs}
\usepackage{ textcomp }
\usepackage{mathtools}
\usepackage{stfloats}

\newcommand{\com}[1]{}

\newcommand{\MTLALL}{MTL\textsubscript{All}}
\newcommand{\newparagraph}[1]{\vspace{0.5em}\noindent\textbf{#1}\hspace{1em}}
\newcommand{\newparagraphsection}[1]{\noindent\textbf{#1}\hspace{1em}}
\newcommand{\shortenfig}{\vspace{-0.3em}}

\usepackage[T1]{fontenc}
\usepackage[utf8]{inputenc}

\usepackage{microtype}

\title{When to Use Multi-Task Learning vs Intermediate Fine-Tuning \\ for Pre-Trained Encoder Transfer Learning}

\author{Orion Weller* \\ Johns Hopkins University
        \And  
        Kevin Seppi \\ Brigham Young University 
        \And
        Matt Gardner \\ Microsoft Semantic Machines }
\begin{document}
\maketitle
\begin{abstract}
Transfer learning (TL) in natural language processing (NLP) has seen a surge of interest in recent years, as pre-trained models have shown an impressive ability to transfer to novel tasks.
Three main strategies have emerged for making use of multiple supervised datasets during fine-tuning: training on an intermediate task before training on the target task (STILTs), using multi-task learning (MTL) to train jointly on a supplementary task and the target task (pairwise MTL), or simply using MTL to train jointly on all available datasets (\MTLALL).
In this work, we compare all three TL methods in a comprehensive analysis on the GLUE dataset suite.
We find that there is a simple heuristic for when to use one of these techniques over the other: pairwise MTL is better than STILTs when the target task has fewer instances than the supporting task and vice versa.
We show that this holds true in more than 92\% of applicable cases on the GLUE dataset and validate this hypothesis with experiments varying dataset size.
The simplicity and effectiveness of this heuristic is surprising and warrants additional exploration by the TL community.
Furthermore, we find that \MTLALL\ is worse than the pairwise methods in almost every case.
We hope this study will aid others as they choose between TL methods for NLP tasks.
\footnote{We make our code publicly available at \url{https://github.com/orionw/MTLvsIFT}.\\ \textasteriskcentered\ Corresponding author, \texttt{oweller2@jhu.edu}}
\end{abstract}

\section{Introduction}

% \begin{figure}[t!]
%     \centering
%     \includegraphics[trim=20 0 0 0,width=0.49\textwidth]{figures/MTLvIFT.pdf}
%     \caption{Overview of the different transfer learning methods: intermediate fine tuning (\textit{STILTs}, \citet{phang2018sentence}) and multi-task learning (\textit{MTL}). STILTs first trains on the supporting task and then uses that trained model to fine-tune again on the target task. In contrast, pairwise MTL trains the contextual encoder on both the target and supporting task simultaneously. What we refer to as \MTLALL\ uses MTL on all $N$ available datasets, where $N > 2$.}
%     \label{fig:flowchart_methods}
% \end{figure}

The standard supervised training paradigm in NLP research is to fine-tune a pre-trained language model on some target task~\cite{peters2018deep,devlin2018bert,raffel2019exploring,gururangan2020don}. When additional non-target supervised datasets are available during fine-tuning, it is not always clear how to best make use of the supporting data \cite{phang2018sentence,phang2020english,liu2019multi,liu2019improving,Pruksachatkun2020IntermediateTaskTL}.  Although there are an exponential number of ways to combine or alternate between the target and supporting tasks, three predominant methods have emerged: (1) fine-tuning on a supporting task and then the target task consecutively, often called STILTs \cite{phang2018sentence}; (2) fine-tuning on a supporting task and the target task simultaneously (here called pairwise multi-task learning, or simply MTL); and (3) fine-tuning on all $N$ available supporting tasks and the target tasks together (\MTLALL, $N > 1$).

Application papers that use these methods generally focus on only one method \cite{Sgaard2017IdentifyingBT,Keskar2019UnifyingQA,Glavas2020IsSS,2019DISCEVALDB,Zhu2019PANLPAM,Weller2020LearningFT,Xu2019DoubleTransferAM,chang2021rethinking}, while a limited amount of papers consider running two. Those that do examine them do so with a limited number of configurations: \citet{phang2018sentence} examines STILTS and one instance of MTL, \citet{changpinyo2018multi,peng2020empirical,schroder2020estimating} compare MTL with \MTLALL, and \citet{wang2018can,talmor2019multiqa,liu2019multi,phang2020english} use \MTLALL\ and STILTs but not pairwise MTL.

In this work we perform comprehensive experiments using all three methods on the 9 datasets in the GLUE benchmark \cite{wang2018glue}. We surprisingly find that a simple size heuristic can be used to determine with more than 92\% accuracy which method to use for a given target and supporting task: when the target dataset is larger than the supporting dataset, STILTS should be used; otherwise, MTL should be used (\MTLALL\ is almost universally the worst of the methods in our experiments).  To confirm the validity of the size heuristic, we additionally perform a targeted experiment varying dataset size for two of the datasets, showing that there is a crossover point in performance between the two methods when the dataset sizes are equal. 
We believe that this analysis will help NLP researchers to make better decisions when choosing a TL method and will open up future research into understanding the cause of this heuristic's success.

\begin{figure*}[t!]
    \centering
    \includegraphics[trim=0 10 0 10,width=1\textwidth]{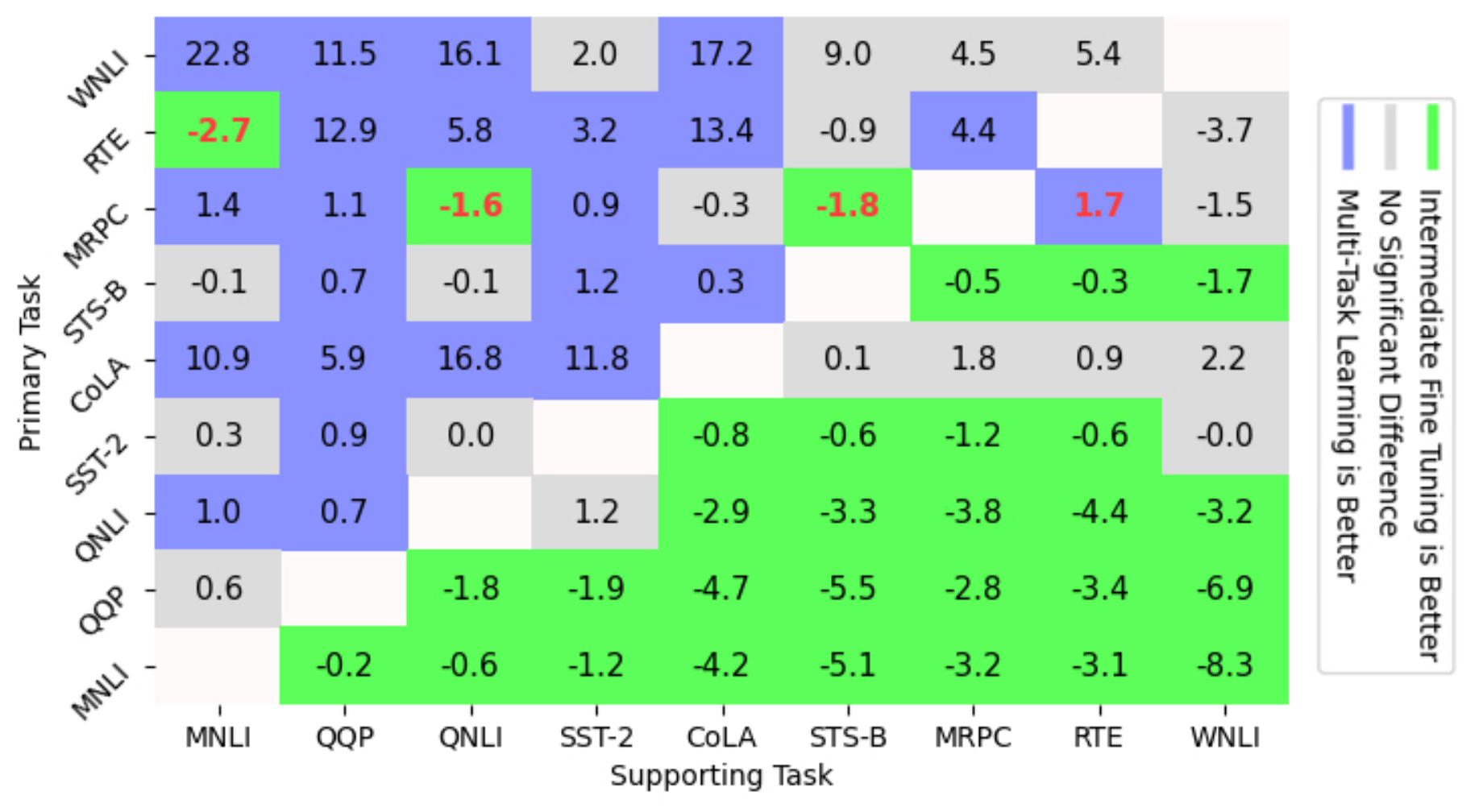}
    \caption{Results comparing intermediate fine tuning (STILTs) vs multi-task learning (MTL). Numbers in cells indicate the absolute percent score difference on the primary task when using MTL instead of STILTs (positive scores mean MTL is better and vice versa). The colors indicate visually the best method, showing a statistically significant difference from the other from using using a two-sided t-test with $\alpha=0.1$. Numbers in red indicate the cells where the size heuristic does not work. Datasets are ordered in descending size (WNLI is the smallest).\shortenfig}
    \label{fig:heatmap}
\end{figure*}

\section{Experimental Settings}
\label{sec:experimental_settings}
\newparagraphsection{Dataset Suite} To conduct this analysis, we chose to employ the GLUE dataset suite, following and comparing to previous work in transfer learning for NLP \cite{phang2018sentence,liu2019multi}.

\newparagraph{Training Framework} We use Huggingface's \textit{transformers} library \cite{wolf2019huggingface} for accessing the pre-trained encoder and for the base training framework. We extend this framework to combine multiple tasks into a single PyTorch \cite{paszke2017automatic} dataloader for MTL and STILTs training. 

Many previous techniques have been proposed for how to best perform MTL \cite{raffel2019exploring,liu2019multi}, but a recent paper by \citet{gottumukkala2020dynamic} compared the main approaches and showed that a new dynamic approach provides the best performance in general. We implement all methods described in their paper and experimented with several approaches (sampling by size, uniformity, etc.). Our initial results found that dynamic sampling was indeed the most effective on pairwise tasks. Thus, for the remainder of this paper, our MTL framework uses dynamic sampling with heterogeneous batch schedules. For consistency, we train the STILTs models using the same code, but include only one task in the dataloader instead of multiple. The \MTLALL\ setup uses the same MTL code, but includes all 9 GLUE tasks.

We train each model on 5 different seeds to control for randomness \cite{dodge2020fine}. For the STILTs method, we train 5 models with different seeds on the supporting task and then choose the best of those models to train with 5 more random seeds on the target task.  For our final reported numbers, we record both the average score and the standard deviation, comparing the MTL approach to the STILTs approach with a two-sample t-test. In total, we train $9*8*5=360$ different MTL versions of our model, 5 \MTLALL\ models, and $9*5+9*5=90$ models in the STILTs setting. 

\newparagraph{Model} We use the DistilRoBERTa model (pre-trained and distributed from the \textit{transformers} library similarly to the DistilBERT model in \citet{sanh2019distilbert}) for our experiments, due to its strong performance and efficiency compared to the full model.  For details regarding model and compute parameters, see Appendix~\ref{app:training}. Our purpose is \textit{not} to train the next state-of-the-art model on the GLUE task and thus the absolute scores are not immediately relevant; our purpose is to show how the different methods score \textit{relative to each other}. We note that we conducted the same analysis in Figure~\ref{fig:heatmap} for BERT and found the same conclusion (see Appendix~\ref{app:bert}), showing that our results extend to other pre-trained transformers.

\begin{figure*}[t!]
    \centering
    \includegraphics[trim=0 10 0 10,width=1\textwidth]{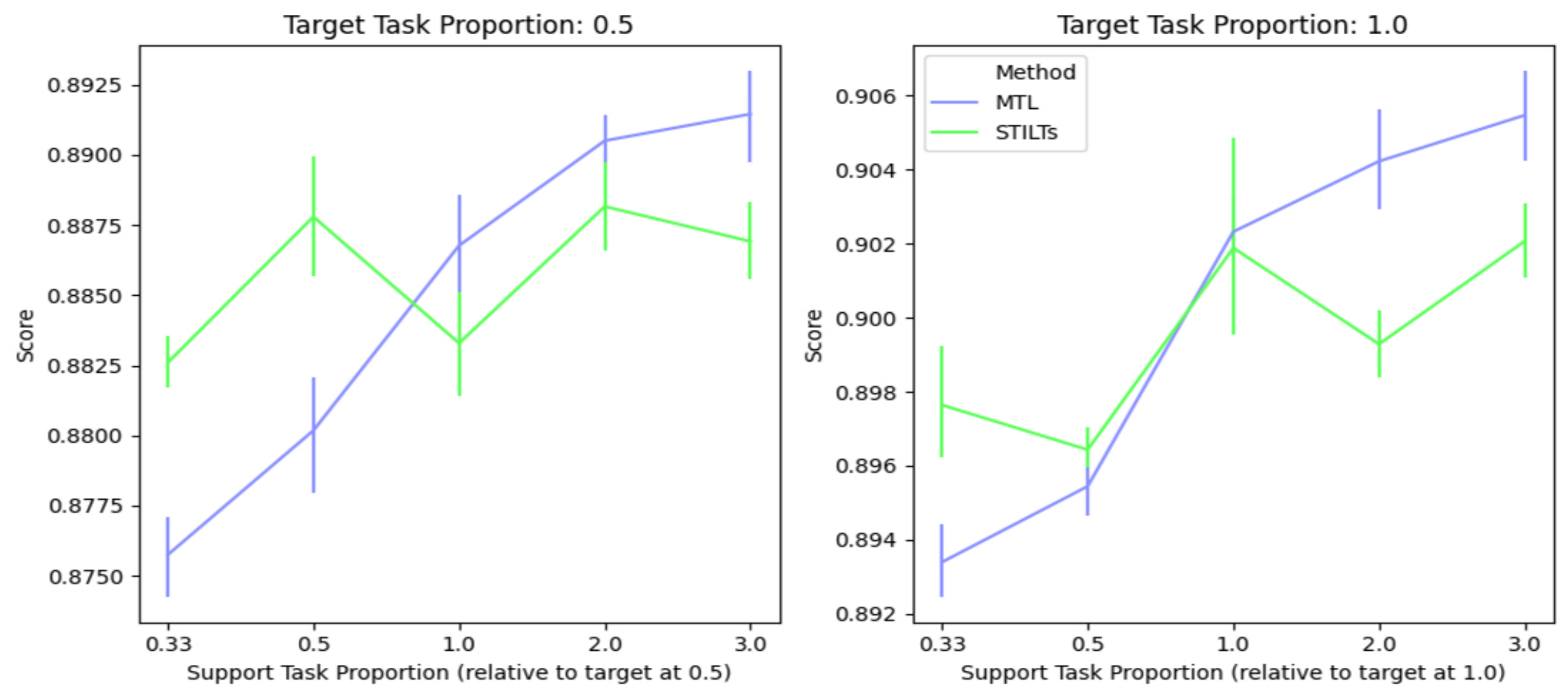}
    \caption{Experiments validating the size heuristic on the (QNLI, MNLI) task pair. The right figure shows training on 100\% of the QNLI training set while the left figure shows training with 50\%. The x-axis indicates the amount of training data of the supporting task (MNLI) relative to the QNLI training set, artificially constrained (e.g. 0.33 indicates that the supporting task is a third of the size of the QNLI training set, etc.). The blue line indicates MTL results while the green line indicates the STILTs method. Error bars indicate a 90\% CI using 5 random seeds.\shortenfig}
    \label{fig:validation}
\end{figure*}

\section{Results}
We provide three different analyses: a comparison of pairwise MTL vs STILTs, experiments varying dataset size to validate our findings, and a comparison of pairwise approaches vs \MTLALL. 

\newparagraph{MTL vs STILTs}
\label{sec:mtl_stilts}
We first calculate the absolute score matrices from computing the MTL and STILTs method on each pair of the GLUE dataset suite, then subtract the STILTs average score matrix from the MTL one (Figure~\ref{fig:heatmap}). Thus, this shows the absolute score gain for using the MTL method instead of the STILTs method (negative scores indicate that the STILTs method was better, etc.). 

However, this matrix does not tell us whether these differences are statistically significant; for this we use a two-sample t-test to compare the mean and standard deviation of each method for a particular cell. 
Scores that are statistically significant are color coded green (if STILTs is better) or blue (if MTL is better), whereas they are coded grey if there is no statistically significant difference. 
We note that although some differences are large (e.g. a 9 point difference on (WNLI, STS-B)) the variance of these results is high enough that there is no statistically significant difference between the STILTs and MTL score distributions.

We order the datasets in Figure~\ref{fig:heatmap} by size, to visually illustrate the trend. The number of green cells in a row is highly correlated with the size of the dataset represented by that row. For example, MNLI is the largest and every cell in the MNLI row is green. QQP is the 2nd largest and every cell in its row is also green, except for (QQP, MNLI). The smallest dataset, WNLI, has zero green cells. 

We can summarize these results with the following size heuristic: \textbf{MTL is better than STILTs when the target task has fewer training instances than the supporting task} and vice versa. In fact, if we use this heuristic to predict which method will be better we find that it predicts 49/53 significant cells, which is equivalent to 92.5\% accuracy. To more clearly visualize which cells it fails to predict accurately, those four cells are indicated with red text. We note that this approach does not hold on the cells that have no statistically significant difference between the two methods: but for almost every significant cell, it does. 

Unfortunately, there is no clear answer to why those four cells are misclassified. Three of the four misclassified cells come when using the MRPC dataset as the target task, but there is no obvious reason why it fails on MRPC.  We recognize that this size heuristic is not an absolute law, but merely a good heuristic that does so with high accuracy: there are still other pieces to this puzzle that this work does not consider, such as dataset similarity.

\newparagraph{Dataset Size Experiments}
\label{sec:synthetic}
In order to validate the size heuristic further we conduct controlled experiments that alter the amount of training data of the supporting task to be above and below the target task. We choose to test QNLI primary with MNLI supporting, as they should be closely related and thus have the potential to disprove this heuristic. We subsample data from the supporting task so that we have a proportion $K$ of the size of the primary task (where $K \in \{1/3, 1/2, 1, 2, 3\}$). By doing so, we examine whether the size heuristic holds while explicitly controlling for the supporting task's size. Other than dataset size, all experimental parameters are the same as in the original comparison (\S\ref{sec:experimental_settings}). 

\begin{table*}[t]
\centering
\small
\begin{tabular}{@{}lrrrrrrrrrr@{}}
\toprule
Approach & Mean & WNLI & STS-B & SST-2 & RTE & QQP & QNLI & MRPC & MNLI & CoLA \\ \midrule
\MTLALL\ & 73.3 & 54.4 & 86.6 & 90.8 & \textbf{67.4} & 80.2 & 84.9 & 85.4 & 74.2 & 35.8 \\
Avg. STILTs & 75.8 & 45.0 & 87.5 & 92.1 & 61.9 & 88.9 & 89.4 & \textbf{87.4} & \textbf{84.0} & 46.4 \\
Avg. MTL & 77.3 & \textbf{56.1} & 87.4 & 91.9 & 66.0 & 85.6 & 87.5 & \textbf{87.4} & 80.8 & \textbf{52.7} \\
Avg. S.H. & \textbf{78.3} & \textbf{56.1} & \textbf{87.7} & \textbf{92.3} & 66.5 & \textbf{89.0} & \textbf{89.6} & 87.3 & \textbf{84.0} & 52.1 \\ \hline
Pairwise Oracle & 80.7 & 57.7 & 88.8 & 92.9 & 76.0 & 89.5 & 90.6 & 90.2 & 84.3 & 56.5 \\
 \bottomrule
\end{tabular}
\caption{Comparison of \MTLALL\ to the pairwise STILTs or MTL approaches. ``S.H" stands for size heuristic. Pairwise Oracle uses the best supplementary task for the given target task using the best pairwise method (STILTs or MTL). All scores are the average of 5 random seeds. We find that on almost every task, pairwise approaches are better than \MTLALL. Bold scores indicate the best score in the column, excluding the oracle.\shortenfig}
\label{tab:mtl_all}
\end{table*}

We also test whether these results hold if the size of the primary dataset is changed (e.g., perhaps there is something special about the current size of the QNLI dataset). We take the same pair and reduce the training set of QNLI in half, varying MNLI around the new number of instances in the QNLI training set as above (e.g. 1/3rd, 1/2, etc.). 

The results of these two experiments are in Figure~\ref{fig:validation}. We can see that as the size of the supporting dataset increases, MTL becomes more effective than STILTs. Furthermore, we find that when both datasets are equal sizes the two methods are statistically similar, as we would expect from the size heuristic (Support Task Proportion$=$1.0). 

Thus, the synthetic experiments corroborate our main finding; the size heuristic holds even on controlled instances where the size of the training sets are artificially manipulated.

\newparagraph{Pairwise TL vs \MTLALL}
\label{sec:mtl_all}
We also experiment with \MTLALL\ on GLUE (see Appendix~\ref{app:sampling} for implementation details). We find that the average pairwise approach consistently outperforms the \MTLALL\ method, except for the RTE task (Table~\ref{tab:mtl_all}) and using the best supporting task outperforms \MTLALL\ in every case (Pairwise Oracle). Thus, although \MTLALL\ is conceptually simple, it is not the best choice w.r.t. the target task score: on a random dataset simply using STILTs or MTL will likely perform better. Furthermore, using the size heuristic on the average supplementary task increases the score by 5 points over \MTLALL\ (78.3 vs 73.3). 

\section{Related Work}
\label{sec:predicting}
A large body of recent work \cite{Sgaard2017IdentifyingBT,vu2020exploring,bettgenhauser2020learning,peng2020empirical,Poth2021WhatTP} exists that examines \textit{when} these transfer learning methods are more effective than simply fine-tuning on the target task. Oftentimes, these explanations involve recognizing catastrophic forgetting \cite{phang2018sentence,pruksachatkun2020intermediate,wang2018can} although recent work has called for them to be re-examined \cite{chang2021rethinking}. This paper is orthogonal to those, as we examine when you should choose MTL or STILTs, rather than when they are more effective than the standard fine-tuning case (in fact, these strategies could be combined to predict transfer and then use the best method). As our task is different, theoretical explanations for how these methods work \textit{in relation to each other} will need to be explored in future work. Potential theories suggested by our results are discussed in Appendix~\ref{app:theory}, and are left to guide those efforts.

\section{Conclusion}
We examined the three main strategies for transfer learning in natural language processing: training on an intermediate supporting task to aid the target task (STILTs), training on the target and supporting task simultaneously (MTL), or training on multiple supporting tasks alongside the target task (\MTLALL).
We provide the first comprehensive comparison between these three methods using the GLUE dataset suite and show that there is a simple rule for when to use one of these techniques over the other. 
This simple heuristic, which holds true in more than 92\% of applicable cases, states that multi-task learning is better than intermediate fine tuning when the target task is smaller than the supporting task and vice versa.  
Additionally, we showed that these pairwise transfer learning techniques outperform the \MTLALL\ approach in almost every case.

\bibliography{anthology,custom}
\bibliographystyle{acl_natbib}

\appendix 
\section{Training and Compute Details}
\label{app:training}
We use the hyperparameters given by the \textit{transformer} library example on GLUE as the default for our model (learning rate of 2e-5, batch size of 128, AdamW optimizer \cite{kingma2014adam}, etc.). We train for 10 epochs, checkpointing every half an epoch and use the best model on the development set for the test set scores. We train on a mix of approximately 10 K80 and P100 GPUs for approximately two weeks for the main experiment, using another week of compute time for the synthetic experiments (\S\ref{sec:synthetic}). Our CPUs use 12-core Intel Haswell (2.3 GHz) processors with 32GB of RAM.  

\begin{table*}[t]
\centering
\small
\begin{tabular}{@{}lrrrrrrrrrr@{}}
\toprule
Approach & Mean & WNLI & STS-B & SST-2 & RTE & QQP & QNLI & MRPC & MNLI & CoLA \\ \midrule
\MTLALL\ Uniform & 63.2 & \textbf{56.1} & 85.1 & 84.0 & 58.3 & 70.4 & 76.4 & 80.3 & 50.7 & 7.8 \\
\MTLALL\ Dynamic & 67.2 & 52.1 & 86.2 & 88.4 & 63.8 & 75.5 & 81.2 & 82.3 & 64.0 & 10.9 \\
\MTLALL\ Size & \textbf{73.3} & 54.4 & \textbf{86.6} & \textbf{90.8} & \textbf{67.4} & \textbf{80.2} & \textbf{84.9} & \textbf{85.4} & \textbf{74.2} & \textbf{35.8} \\
 \hline
Avg. STILTs & 75.8 & 45.0 & 87.5 & 92.1 & 61.9 & 88.9 & 89.4 & \textbf{87.4} & \textbf{84.0} & 46.4 \\
Avg. MTL & 77.3 & \textbf{56.1} & 87.4 & 91.9 & 66.0 & 85.6 & 87.5 & \textbf{87.4} & 80.8 & \textbf{52.7} \\
Avg. S.H. & \textbf{78.3} & \textbf{56.1} & \textbf{87.7} & \textbf{92.3} & 66.5 & \textbf{89.0} & \textbf{89.6} & 87.3 & \textbf{84.0} & 52.1 \\
\hline
Pairwise Oracle & \textbf{80.7} & \textbf{57.7} & \textbf{88.8} & \textbf{92.9} & \textbf{76.0} & \textbf{89.5} & \textbf{90.6} & \textbf{90.2} & \textbf{84.3} & \textbf{56.5} \\
 \bottomrule
\end{tabular}
\caption{Comparison of \MTLALL\ to the pairwise STILTs or MTL approaches. ``S.H" stands for size heuristic. Pairwise Oracle uses the best supplementary task for the given target task using the best pairwise method (STILTs or MTL). All scores are the average of 5 random seeds. Note that \MTLALL\ was run with three different sampling methods (top half). We find that on almost every task, pairwise approaches are better than \MTLALL. Bold scores indicate the best score in the column for the given section.}
\label{tab:app_mtl_all}
\end{table*}

\section{Pairwise Approaches vs \MTLALL}
\label{app:sampling}
\paragraph{Experimental Setup}
We use \MTLALL\ with three different sampling methods: uniform sampling, sampling by dataset size, and dynamic sampling. To illustrate the difference between \MTLALL\ and the pairwise methods, we show the average score across all supplementary tasks for MTL and STILTs. We also show the average score found by choosing MTL or STILTs using the size heuristic as \textit{Ave. S.H.}.  Finally, we report the score from the best task using the best pairwise method, which we call the \textit{Pairwise Oracle}. The results are shown in Table~\ref{tab:app_mtl_all}.

\paragraph{Results} Although dynamic sampling was more effective for the pairwise tasks, we find that dynamic sampling was worse than sampling by size when using MTL on all nine datasets (top half of Table~\ref{tab:app_mtl_all}).

However, when the \MTLALL\ method is compared to the pairwise methods, it does not perform as well (bottom half of Table~\ref{tab:app_mtl_all}). We see that the Pairwise Oracle, which uses the best supplementary task for the given target task, outperforms all methods by a large margin. Thus, although \MTLALL\ is conceptually simple, it is not the best choice with respect to target task accuracy. Furthermore, if you could predict which supplementary task would be most effective (Pairwise Oracle, c.f. Section~\ref{sec:predicting}, \citet{vu2020exploring,Poth2021WhatTP}, etc.), you would be able to make even larger gains over \MTLALL. 

\section{Theories for Transfer Effectiveness}
\label{app:theory}
Previous work often invokes ideas such as catastrophic forgetting to describe why STILTs or MTL does or does not improve over the basic fine-tuning case \cite{phang2018sentence,pruksachatkun2020intermediate,wang2018can}. However, as our work provides a novel comparison of MTL vs. STILTs there exists no previous work that shows how these methods differ in any practical or theoretical terms (e.g. does MTL or STILTs cause more catastrophic forgetting of the target task). Furthermore, previous explanations for why the STILTs method works has been called into question \cite{chang2021rethinking}, leaving it an open research area. 

A naive explanation for our task would be to think that when the target task is larger, STILTs should be worse because of catastrophic forgetting, whereas MTL would still have access to the supporting task. However, for STILTs this catastrophic forgetting would mainly effect the supporting task performance, not the target task performance, making that explanation unlikely in some contexts (e.g. when the tasks are not closely related). One potential explanation based on our results is that a small supporting task is best used to provide a good initialization for a larger target task (e.g. STILTs) while a large supporting task used for initialization would change the weights too much for the small target task to use effectively (thus making MTL the more effective strategy for a larger supporting task). Another explanation could be that a larger target task does not benefit from MTL (and perhaps is harmed by it, e.g. catastrophic interference) and therefore, STILTs is more effective - while MTL is more effective for small target tasks. However, all of these explanations also fail to take into account task relatedness, which likely also plays a role in the theoretical explanation (although even that too, has been called into question with \citet{chang2021rethinking}).

We thus note that there are a myriad of possible explanations (and the answer is likely a complex combination of possible explanations), but these are out of the scope of this work. Our work aims to show what happens in practice, rather than proposing a theoretical framework. As theoretical explanations for transfer learning are still an active area of research, we leave them to future work and provide this empirical comparison to guide their efforts and the current efforts of NLP researchers and practitioners.

\section{Alternate Model: BERT}
\label{app:bert}
\begin{figure*}[t!]
    \centering
    \includegraphics[trim=0 10 0 10,width=1\textwidth]{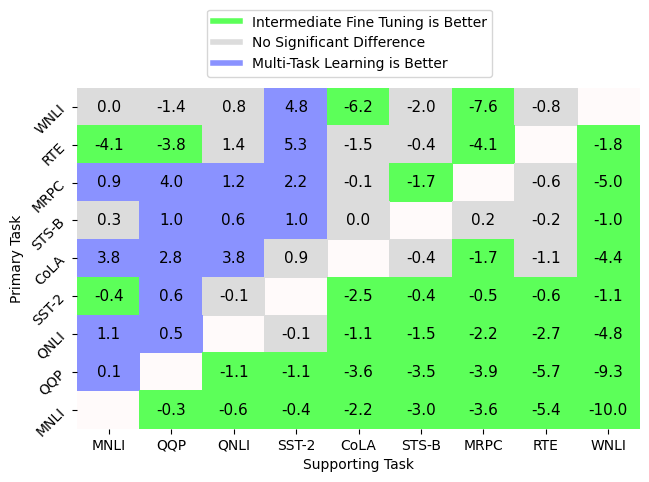}
    \caption{Results comparing intermediate fine tuning (STILTs) vs multi-task learning (MTL) with the BERT model. Numbers in cells indicate the absolute percent score difference on the primary task when using MTL instead of STILTs (positive scores mean MTL is better and vice versa). The colors indicate visually the best method, showing a statistically significant difference from the other from using using a two-sided t-test with $\alpha=0.1$. Datasets are ordered in descending size.\shortenfig}
    \label{fig:bert}
\end{figure*}

We conduct the same analysis as Figure~\ref{fig:heatmap} with the BERT model and find similar results (Figure~\ref{fig:bert}, thus showing that our results transfer to other pretrained transformer models. We follow previous work in using two different pre-trained models for our analysis \cite{talmor2019multiqa,phang2018sentence}.

\section{Additional Background Discussion}
\label{app:discussion}
In this section we will show how the size heuristic is supported by and helps explain the results of previous work in this area. \textbf{Although this section is not crucial to the main result of our work, we include it to help readers who may not be as familiar with the related work}. We examine two works in depth and then discuss broader themes of related work.

\begin{table}[]
\centering
\begin{tabular}{@{}lr@{}}
\toprule
Model & RTE accuracy \\ \midrule
GPT $\to$ RTE & 54.2 \\
GPT $\to$ MNLI $\to$ RTE & \textbf{70.4} \\
GPT $\to$ \{MNLI, RTE\}  & 68.6 \\
GPT $\to$ \{MNLI, RTE\} $\to$ RTE & 67.5 \\
 \bottomrule
\end{tabular}
\caption{Table reproduced from \citet{phang2018sentence}. Their comparison of STILTs against MTL setups for GPT, with MNLI as the intermediate task and RTE as the target task. Only one run was reported (e.g. no standard error or confidence intervals).}
\label{tab:stilts}
\end{table}

\paragraph{BERT on STILTs \citet{phang2018sentence}}
\label{sec:stilts}
This work defined the acronym STILTs, or \textit{Supplementary Training on Intermediate Labeled-data Tasks}, which has been an influential idea in the community \cite{Voskarides2019ILPSAT,Yan2020SQLGV,Clark2020ELECTRAPT}. To determine the effect of the intermediate training, the authors computed the STILTs matrix of each pair in the GLUE dataset. As our model and training framework are different from their methodology, we cannot compare our matrix with the absolute numbers in their matrix. However, at the end of Section 4 in their paper, they conduct an experiment with MTL and compare the results to their STILTs matrix (their experimental results are reproduced in Table~\ref{tab:stilts} for convenience). Their analysis uses MNLI as the supporting task and RTE as the target task, trying MTL, STILTs, MTL+fine-tuning, and only fine-tuning on RTE. Their results show that STILTs provides the highest score, with all MTL varieties being worse.  From this they conclude that MTL is worse than STILTs.

\textit{How does this compare to our results?} In Figure~\ref{fig:heatmap} we see that our results also show that the STILTs method is better than the MTL method for the (RTE, MNLI) pair, showing that our results are consistent with those in the literature. Furthermore, we find that this is one of the 4 significant cells in our matrix where the size heuristic does not accurately predict the best method. It is unfortunate that the task they decided to pick happened to be one of the anomalies. Thus, our paper extends and completes their results with more rigor.

\paragraph{MultiQA \citet{talmor2019multiqa}}
\label{app:multiqa}
MultiQA showed that using MTL on a variety of question-answering (QA) datasets made it possible to train a model that could outperform the current SOTA on those QA datasets. They used an interesting approach to MTL, pulling 15k examples from each of the 5 major datasets to compose one new ``MTL" task, called Multi-75K.  They then show results for STILTs transfer on those same datasets along with the MTL dataset (their data is reproduced with new emphasis in Appendix~\ref{app:multiqa} Table~\ref{tab:multiqa} for convenience). We note that this STILTs-like transfer with the ``MTL" dataset is an equivalent method to doing MTL and then fine-tuning on the target task, reminiscent of the third example in \citet{phang2018sentence} (Table~\ref{tab:stilts}, GPT$\,\to\,$\{MNLI, RTE\}$\,\to\,$RTE, c.f. Appendix~\ref{app:ft_mtl}). 

\textit{How does this relate to our results?} The size heuristic says that MTL is better than STILTs when the target task has fewer training instances. In the MultiQA paper the size of each training set is artificially controlled to be the same number (75k instances), thus our size heuristic would say that the methods should be comparable. Although no error bounds or standard deviations are reported in their paper (which makes the exact comparison difficult), we see that the MTL approach performs equal or better on almost half of the datasets. Thus, although the MultiQA paper is not strictly comparable to our work due to their training setup (the MTL+fine tuning), their results agree with our hypothesis as well.

For convenience, Table 4 from \citet{talmor2019multiqa} is reproduced here in the appendix. The top half contains the results using the DocQA model while the bottom half uses BERT. Note that both model's Multi-75K scores perform approximately similar to the STILTs methods, which is expected given that they are the same size. TQA-G and TQA-W come from the same dataset. As stated in the body of this paper, no standard deviation is reported in the MultiQA paper and thus it is hard to know whether the difference in results are statistically significant. Even if all results were statistically significant, which is highly unlikely, each of the Multi-75K models perform equal or better on 2 of the 6 tasks, which is not statistically different from random.

\begin{table*}[]
\centering
\begin{tabular}{@{}lllllll@{}}
\toprule
          & SQuAD & NewsQA & SearchQA & TQA-G & TQA-W & HotpotQA \\ \midrule
SQuAD     & -     & \textbf{33.3}   & 39.2     & 49.2  & 34.5  & 17.8     \\
NewsQA    & 59.6  & -      & 41.6     & 44.2  & 33.9  & 16.5     \\
SearchQA  & 57    & 31.4   & -        & \textbf{57.5}  & 39.6  & \textbf{19.2}     \\
TQA-G     & 57.7  & 31.8   & \textbf{49.5}     & -     & \textbf{41.4}  & 19.1     \\
TQA-W     & 57.6  & 31.7   & 44.4     & 50.7  & -     & 17.2     \\
HotpotQA  & \textbf{59.8}  & 32.4   & 46.3     & 54.6  & 37.4  & -        \\
Multi-75K & \textbf{59.8}  & 33.0     & 47.5     & 56.4  & 40.4  & \textbf{19.2}     \\
\midrule
SQuAD     & -     & 41.2   & 47.8     & 55.2  & 45.4  & 20.8     \\
NewsQA    & \textbf{72.1}  & -      & 47.4     & 55.9  & 45.2  & 20.6     \\
SearchQA  & 70.2  & 40.2   & -        & \textbf{57.3}  & 45.5  & 20.4     \\
TQA-G     & 69.9  & 41.2   & \textbf{50.0}       & -     & 46.2  & 20.8     \\
TQA-W     & 71.0    & 39.2   & 48.4     & 55.7  & -     & \textbf{20.9}     \\
HotpotQA  & 71.2  & 39.5   & 48.6     & 56.6  & 45.6  & -        \\
Multi-75K & 71.5  & \textbf{42.1}   & 48.5     & 56.6  & \textbf{46.5}  & 20.4     \\ \bottomrule
\end{tabular}
\caption{Results taken from the right half of Table 4 in the MultiQA paper \cite{talmor2019multiqa} as that section is directly relevant to this work (the \textit{self} row containing only standard fine-tuning is removed for clarity). Emphasis changed to reflect the best score in the model's column instead of the best non-MTL score.\label{tab:multiqa}}
\end{table*}

\paragraph{Combining All Tasks}
Our results using \MTLALL\ showed that although \MTLALL\ is conceptually easy (just put all the datasets together) it does not lead to the best performance. We find similar results in \citet{wang2018can}, where in their Table 3 they show that the STILTs approach outperforms the \MTLALL\ approach for all but one task. Additionally, in the follow up work from the initial STILTs paper \cite{phang2020english} they find that although \MTLALL\ has a slightly higher average performance in the cross-lingual setting, it is worse than the pairwise approach in 75\% of the evaluated tasks.

The current literature (and our work) seems to suggest that naively combining as many tasks as possible may not be the best approach. However, more work is needed to understand the training dynamics of \MTLALL. 

\paragraph{Combining Helpful Tasks} In this paper, we only examine the difference between pairwise MTL, STILTs or \MTLALL, due to time and space. Although it is possible that our heuristic may extrapolate to transfer learning with more than two tasks, computing the power set of the possible task combinations for MTL and STILTs would be extremely time and resource intensive. We leave it to future work to examine how the size heuristic may hold when using more than two datasets at a time.

Additionally, there may be further value in computing this power set: \citet{changpinyo2018multi} showed that taking the pairwise tasks that proved beneficial in pairwise MTL and combining them into a larger MTL set (an ``Oracle" set) oftentimes provides higher scores than pairwise MTL. Exploring which subsets of tasks provide the best transfer with which method would be valuable future work.

\paragraph{Dataset Size in TL} Dataset size has been used often in transfer learning techniques \cite{Sgaard2017IdentifyingBT,Pruksachatkun2020IntermediateTaskTL,Poth2021WhatTP}. Our size heuristic, although related, focuses on a different problem: whether to use MTL or STILTs. Thus, our work provides additional insight into how the size of the dataset is important for transfer learning.

\paragraph{Fine-tuning after MTL}
\label{app:ft_mtl}
Many papers that use \MTLALL\ also perform some sort of fine-tuning after the MTL phase. Since fine-tuning after MTL makes the MTL phase an intermediate step, it essential combines the STILTs and MTL methods into a single STILTs-like method.  However, whether fine-tuning after MTL is better than simply MTL is still controversial: for example, \citet{liu2019multi}, \citet{raffel2019exploring}, and \citet{talmor2019multiqa} say that fine-tuning after MTL helps but \citet{lourie2021unicorn} and \citet{phang2018sentence} say that it doesn't. However, \citet{raffel2019exploring} is the only one whose experiments include multiple random seeds, giving more credence to their results. However, due to the difference of opinion it is unclear which method is actually better; we leave this to future work. 

\section{GLUE Dataset Sizes and References}
To give credit to the original authors and to provide the exact sizes, we provide Table~\ref{tab:sizes}.

\begin{table}[t!]
\small
\centering
\begin{tabular}{@{}lp{3.5cm}r@{}}
\toprule
Dataset  & Citation & Training Size \\ \midrule
MNLI        &    \citet{williams2018broad}  & 392,662      \\
QQP         &    No citation, \hyperlink{https://data.quora.com/First-Quora-Dataset-Release-Question-Pairs}{link here}  & 363,846       \\
QNLI        &     \citet{levesque2011winograd}   & 104,743      \\
SST-2          &    \citet{socher2013recursive}    & 67,349    \\
CoLA             &    \citet{warstadt2018neural}  & 8,551    \\
STS-B           &     \citet{cer-etal-2017-semeval}   & 5,749    \\
MRPC            &    \citet{dolan2005automatically}   & 3,668     \\
RTE           &    \citet{dagan2006pascal}*   & 2,490    \\
WNLI             &     \citet{levesque2011winograd}  & 635    \\
\bottomrule
\end{tabular}
\caption{Sizes of the datasets in GLUE \cite{wang2018glue} in descending order, along with their original citations. RTE is compiled from these sources: \citet{dagan2006pascal,bar2006second,giampiccolo2007third,bentivogli2009fifth}\label{tab:sizes}}
\end{table}

\end{document}